\documentclass[runningheads]{llncs}
\usepackage{graphicx}
\usepackage{comment}
\usepackage{amsmath,amssymb} 
\usepackage{color}

\usepackage{latexsym}
\usepackage{amsmath}
\usepackage{booktabs} 
\usepackage{subfigure}
\usepackage{multirow}

\begin{document}
\pagestyle{headings}
\mainmatter
\def\ECCVSubNumber{5757}  

\title{Structure-Aware Generation Network for Recipe Generation from Images} 

\titlerunning{Structure-Aware Generation Network for Recipe Generation from Images}
%
\author{Hao Wang\inst{1,2} \and
Guosheng Lin\inst{1} \and
Steven C. H. Hoi\inst{3} \and
Chunyan Miao \inst{1,2}\thanks{Corresponding author} }
\authorrunning{Wang et al.}

\institute{School of Computer Science and Engineering, \\
Nanyang Technological University, Singapore \\
\and
Joint NTU-UBC Research Centre of Excellence \\
in Active Living for the Elderly, NTU
\and
Singapore Management University\\
\email{\{hao005,gslin,ascymiao\}@ntu.edu.sg}\quad \email{chhoi@smu.edu.sg}}
\maketitle

\begin{abstract}
Sharing food has become very popular with the development of social media. For many real-world applications, people are keen to know the underlying recipes of a food item. In this paper, we are interested in automatically generating cooking instructions for food. We investigate an open research task of generating cooking instructions based on only food images and ingredients, which is similar to the image captioning task. However, compared with image captioning datasets, the target recipes are long-length paragraphs and do not have annotations on structure information. To address the above limitations, we propose a novel framework of Structure-aware Generation Network (SGN) to tackle the food recipe generation task. Our approach brings together several novel ideas in a systematic framework: (1) exploiting an unsupervised learning approach to obtain the sentence-level tree structure labels before training; (2) generating trees of target recipes from images with the supervision of tree structure labels learned from (1); and (3) integrating the inferred tree structures with the recipe generation procedure. Our proposed model can produce high-quality and coherent recipes, and achieve the state-of-the-art performance on the benchmark Recipe1M dataset.
\end{abstract}

\section{Introduction}

Food-related research with the newly evolved deep learning-based techniques is becoming a popular topic, as food is essential to human life. One of the important and challenging tasks under the food research domain is recipe generation \cite{salvador2019inverse}, where we are producing the corresponding and coherent cooking instructions for specific food. 

In the recipe generation dataset Recipe1M \cite{salvador2017learning}, we generate the recipes conditioned on food images and ingredients. The general task setting of recipe generation is almost the same as that of image captioning \cite{chen2015microsoft}. Both of them target generating a description of an image by deep models.  However, there still exist two big differences between recipe generation and image captioning: (i) the target caption length and (ii) annotations on structural information. 

First, most popular image captioning datasets, such as Flickr \cite{plummer2015flickr30k} and MS-COCO dataset \cite{chen2015microsoft}, only have one sentence per caption. By contrast, cooking instructions are paragraphs, containing multiple sentences to guide the cooking process, which cannot be fully shown in single food image. Although Recipe1M has ingredient information, the ingredients are actually mixed in cooked food images. Therefore, generating lengthy recipes using traditional image captioning model may hardly capture the whole cooking procedure. Second, the lack of structural information labeling is another challenge in recipe generation. For example, MS-COCO has precise bounding box annotations in images, giving scene graph information for caption generation. This structural information provided by the official dataset makes it easier to recognize the objects, their attributes and relationships within an image. While in food images, different ingredients are mixed when cooked. Hence, it is difficult to obtain the detection labeling for food images. 

\begin{figure*}[t]
\begin{center}
\includegraphics[width=\textwidth]{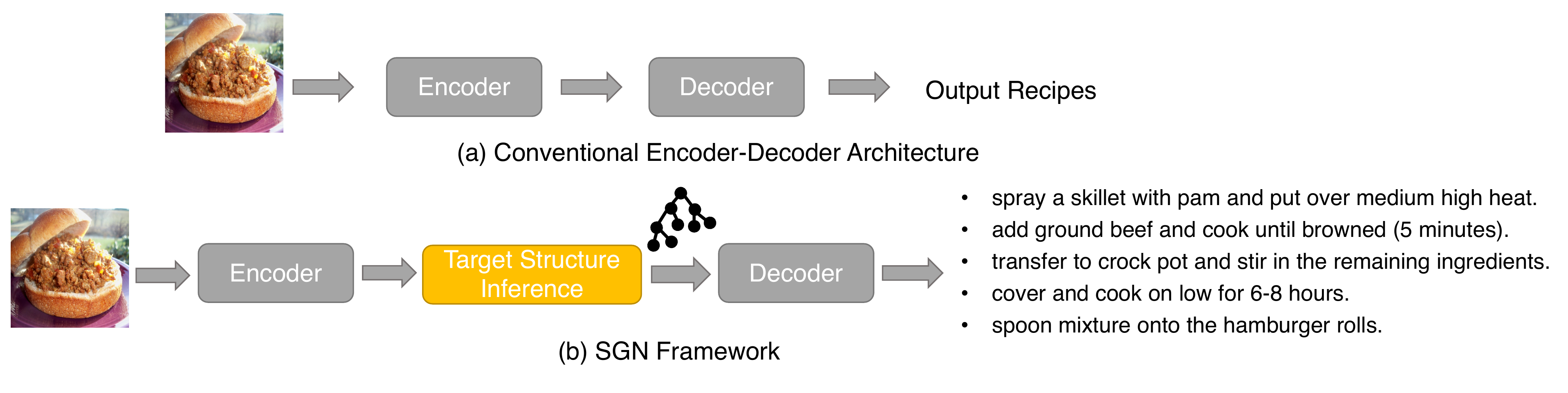}
\end{center}
   \caption{Comparison between conventional image captioning model and our proposed Structure-aware Generation Network (SGN) for recipe generation. Before generating target recipes, we infer the tree structures of recipes first, then we use graph attention networks to give tree embeddings. Based on the structure information, we can generate better recipes.}
\label{fig:demo}
\end{figure*}

Benefiting from recent advances in language parsing, some research, such as ON-LSTM \cite{shen2018ordered}, utilizes an unsupervised way to produce word-level parsing trees of sentences and achieve good results. Inspired by that, we extend the ON-LSTM architecture to do sentence-level tree structure generation. We propose to train the extended ON-LSTM with quick thoughts manner \cite{logeswaran2018efficient}, to capture the order information inside recipes. By doing so, we get the recipe tree structure labels.

After we obtained the recipe structure information, we propose a novel framework named \textbf{S}tructure-aware \textbf{G}eneration \textbf{N}etwork (SGN) to integrate the tree structure information with the training and inference phases. SGN is implemented to add a target structure inference module on the recipe generation process. Specifically, we propose to use a RNN to generate the recipe tree structures from food images. Based on the generated trees, we adopt the graph attention networks to embed the trees, in an attempt to giving the model more guidance when generating recipes. With the tree structure embeddings, we make the generated recipes remain long-length as the ground truth, and improve the generation performance considerably.

Our contributions can be summarized as:
\begin{itemize}
   \item We propose a recipe2tree module to capture sentence-level tree structures for recipes, which is adopted to supervise the following img2tree module. We use an unsupervised approach to learn latent tree structures. 
   \item We propose to use the img2tree module to generate recipe tree structures from food images, where we adopt a RNN for conditional tree generation.
   \item We propose to utilize the tree2recipe module, which encodes the inferred tree structures. It is implemented with graph attention networks, and boosts the recipe generation performance.
\end{itemize}

Figure \ref{fig:demo} shows a comparison between vanilla image captioning model and our proposed SGN. SGN outperforms state-of-the-art baselines for recipe generation on Recipe1M dataset \cite{salvador2017learning} significantly. We conduct extensive experiments to evaluate performance of SGN. Finally, we present qualitative results of our proposed methods, and visualizations of the generated recipe results.

\section{Related Work}

\subsection{Image captioning}
Image captioning task is defined as generating the corresponding text descriptions from images. Based on MS-COCO dataset \cite{chen2015microsoft}, most existing image captioning techniques adopt deep learning-based model. One popular approach is Encoder-Decoder architecture \cite{vinyals2015show,bahdanau2014neural,golland2010game,kazemzadeh2014referitgame}, where a CNN is used to obtain the image features along with object detection, then a language model is used to convert the image features into text. 

Since image features are fed only at the beginning stage of generation process, the language model may face vanishing gradient problem \cite{hossain2019comprehensive}. Therefore, image captioning model is facing challenges in long sentence generation \cite{bahdanau2014neural}. To enhance text generation process, \cite{chen2019counterfactual,gu2019unpaired} involve scene graph into the framework. However, scene graph generation rely heavily on object bounding box labeling, which is provided by MS-COCO dataset. When we shift to some other datasets without rich annotation, we can hardly obtain the graph structure information of the target text. Meanwhile, crowdsourcing annotation is high-cost and may not be reliable. Therefore, we propose to produce tree structures for paragraphs unsupervisedly, helping the recipe generation task in Recipe1M dataset \cite{salvador2017learning}. 

\subsection{Multimodal Food Computing}
Food computing \cite{min2019survey} has raised great interest recently, it targets applying computational approaches for analyzing multimodal food data for recognition \cite{bossard2014food}, retrieval \cite{salvador2017learning,wang2019learning,wang2020cross} and generation \cite{salvador2019inverse} of food. Recipe generation is a challenging task, it is mainly because that cooking instructions (recipes) contain more than one sentence. Besides, the input images of food datasets \cite{salvador2017learning,zhou2018towards,chandu2019storyboarding} suffer from quality issues. Among the existing food datasets, YouCook2 \cite{zhou2018towards} and Storyboarding \cite{chandu2019storyboarding} use image sequence as input, which is not a general image captioning task setting. Recipe1M \cite{salvador2017learning,salvador2019inverse} contains static food image, ingredient and cooking instruction information for each food sample. Therefore, we choose Recipe1M \cite{salvador2017learning} to evaluate our proposed methods.

\subsection{Language Parsing}
Parsing is served as one effective language analysis tool, it can output the tree structure of a string of symbols. Generally, language parsing is divided into word-level and sentence-level parsing. Word-level parsing is also known as grammar induction, which aims at learning the syntactic tree structure from corpora data. 
Especially, Shen et al. \cite{shen2018ordered} propose to use ON-LSTM, which equips the LSTM architecture with an inductive bias towards learning latent tree structures. They train the model with normal language modeling way, at the same time they can get the parsing output induced by the model.

Sentence-level parsing is used to identify the elementary discourse units in a text, and it brings some benefits to discourse analysis. Many recent works attempted to use complex model with labeled data to achieve the goal \cite{jia2018modeling,lin2019unified}. Here we extend ON-LSTM \cite{shen2018ordered} for unsupervised sentence-level parsing, which is trained using quick thoughts \cite{logeswaran2018efficient}.

\section{Method}
\begin{figure*}[t]
\begin{center}
\includegraphics[width=\textwidth]{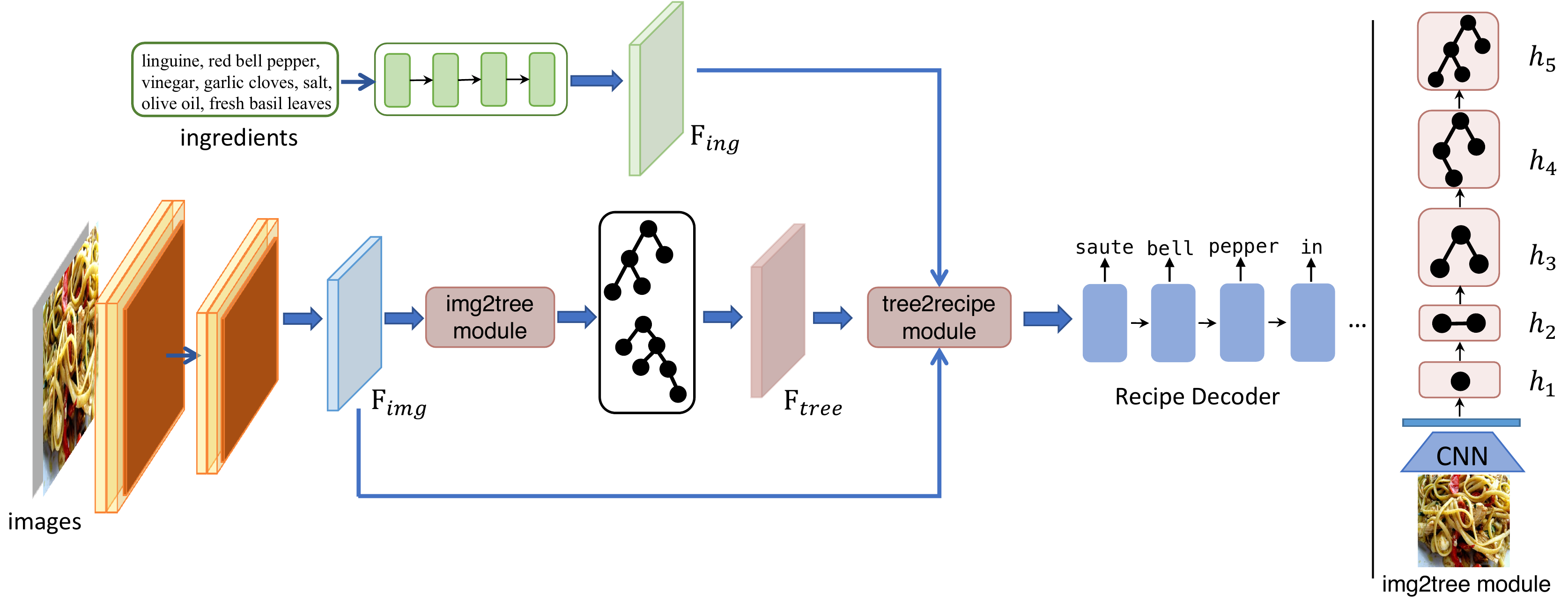}
\end{center}
   \caption{\textbf{Left:} our proposed framework for effective recipe generation. \textbf{Right:} the img2tree module. The ingredients and food images are embedded by a pretrained language model and CNN respectively to produce the output features $\mathrm{F}_{ing}$ and $\mathrm{F}_{img}$. Before language generation, we first infer the tree structure of target cooking instructions. To do so, we utilize the img2tree module, where a RNN produces the nodes and edge links step-by-step based on $\mathrm{F}_{img}$. Then in tree2recipe module, we adopt graph attention networks (GAT) to encode the generated tree adjacency matrix, and get the tree embedding $\mathrm{F}_{tree}$. We combine $\mathrm{F}_{ing}$, $\mathrm{F}_{img}$ and $\mathrm{F}_{tree}$ to construct a final embedding for recipe generation, which is performed using a transformer.}
\label{fig:pipeline}
\end{figure*}

We now present our proposed model SGN for recipe generation from food images. We show the framework of our proposed model in Figure \ref{fig:pipeline}.

\subsection{Overview}
Given the food images and ingredients, our goal is to generate the cooking instructions. Different from image captioning task in MS-COCO \cite{chen2015microsoft,luo2018discriminability}, where the target caption only have one sentence, the cooking instruction is a paragraph, containing more than one sentence, and the maximum sentence number in Recipe1M dataset \cite{salvador2017learning} is $19$. If we infer the recipes directly from the images, i.e. use a decoder conditioned on image features for generation \cite{salvador2019inverse}, it is difficult for model to fully capture the structured cooking steps. That may result in the generated paragraphs incomplete. Hence, we believe it necessary to infer paragraph structure during recipe generation phase. 

\begin{figure*}[t]
\begin{center}
\includegraphics[width=0.8\textwidth]{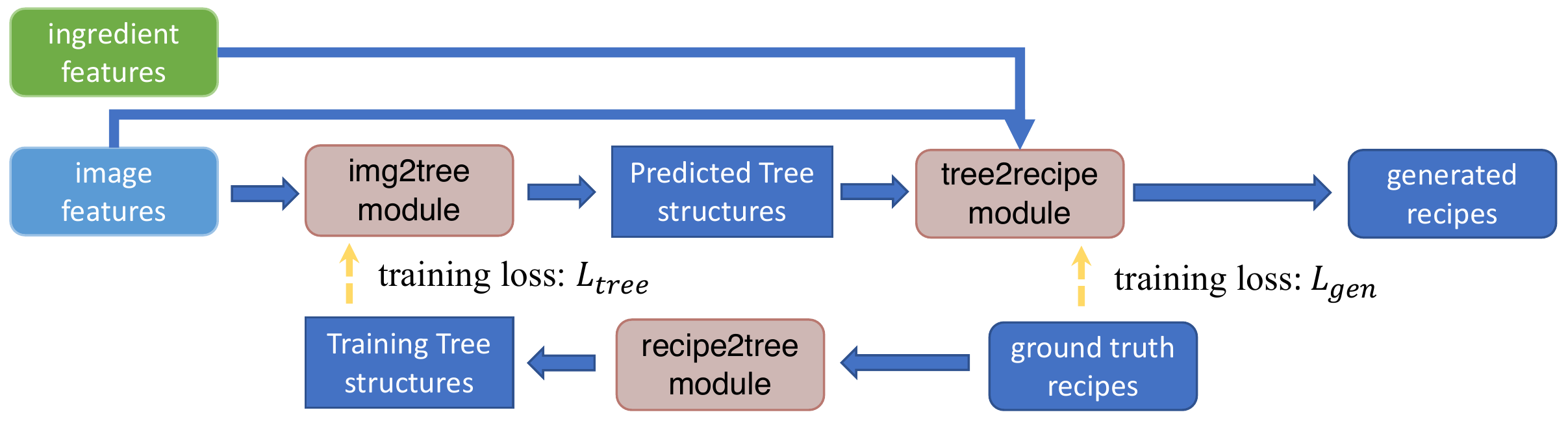}
\end{center}
   \caption{The concise training flow of our proposed SGN.}
\label{fig:training}
\end{figure*}

To infer the sentence-level tree structure from food images, we need labels to supervise the tree generation process. However, in Recipe1M dataset \cite{salvador2017learning}, there has no paragraph tree structure labeling for cooking instructions. And it is very time-consuming and unreliable to use crowdsourcing to give labels. Therefore, in the first step, we use the proposed recipe2tree module to produce the tree structure labels with an unsupervised way. Technically, we use hierarchical ON-LSTM \cite{shen2018ordered} to encode the cooking instructions and train the ON-LSTM with quick thoughts approach \cite{logeswaran2018efficient}. Then we can obtain the latent tree structure of cooking instructions.

During training phase, we input food images and ingredients to our proposed model. We try two different language models to encode ingredients, i.e. non-pretrained and pretrained model, to get the ingredient features $\mathrm{F}_{ing}$. In non-pretrained model training, we use one word embedding layer to give $\mathrm{F}_{ing}$. Besides, we adopt BERT \cite{devlin2018bert} for ingredient embedding, which is one of the state-of-the-art NLP pretrained models. 

In image process branch, we adopt a CNN to encode the food images and get the image features $\mathrm{F}_{img}$. Based on $\mathrm{F}_{img}$, we generate the trees and make them align with that produced by the recipe2tree module. Specifically, we transform the tree structure to a $1$-dimensional adjacency sequence for RNN to generate, where the RNN's initial state is image feature $\mathrm{F}_{img}$. To incorporate the generated tree structure into the recipe generation process, we get the tree embedding $\mathrm{F}_{tree}$ with graph attention networks (GATs) \cite{velivckovic2017graph}, and concatenate it with the image features $\mathrm{F}_{img}$ and ingredient features $\mathrm{F}_{ing}$. We then generate the recipes conditioned on the joint features $\langle \mathrm{F}_{tree}, \mathrm{F}_{img}, \mathrm{F}_{ing} \rangle$ with a transformer \cite{vaswani2017attention}. 

Our proposed framework is optimized over two objectives: to generate reasonable recipes given the food images and ingredients; and to produce the sentence-level tree structures of target recipes. The overall
objective is given as:
\begin{equation}
    L = \lambda_1L_{gen} + \lambda_2L_{tree}
\end{equation}
where $\lambda_1$ and $\lambda_2$ are trade-off parameters. $L_{gen}$ controls the recipe generation training with the input of $\langle \mathrm{F}_{tree}, \mathrm{F}_{img}, \mathrm{F}_{ing} \rangle$, and outputs the probabilities of word tokens. $L_{tree}$ is the tree generation loss, supervising the img2tree to generate trees from images. The training flow in shown in Figure \ref{fig:training}. 

\subsection{ON-LSTM Revisit}
Ordered Neurons LSTM (ON-LSTM) \cite{shen2018ordered} is proposed to infer the underlying tree-like structure of language while learning the word representation. It can achieve good performance in unsupervised parsing task. ON-LSTM is constructed based on the intuition that each node in the tree can be represented by a set of neurons in the hidden states of recurrent neural networks. To this end, ordered neurons is a inductive bias, where high-ranking neurons store long-term information, while low-ranking neurons contain short-term information that can be rapidly forgotten. Instead of acting independently on each neuron, the gates of ON-LSTM are dependent on the others by enforcing the order in which neurons should be updated. Hence, ON-LSTM is able to discern a hierarchy of information between neurons. With the learnt ranking, the top-down greedy parsing algorithm \cite{shen2017neural} is used for unsupervised constituency parsing. However, ON-LSTM is originally trained by language modeling way and learns the word-level order information. To unsupervisedly produce sentence-level tree structure, we extend ON-LSTM in recipe2tree module.

\subsection{Recipe2tree Module}
In recipe2tree module, we propose to use hierarchical ON-LSTM, i.e. word-level and sentence-level ON-LSTM, to train Recipe1M data. Specifically, in word-level ON-LSTM, we input the word tokens and use the output features as the sentence embeddings. The sentence embeddings will be fed into the sentence-level ON-LSTM for end-to-end training. 

Since the original training way \cite{shen2018ordered}, such as language modeling or seq2seq \cite{britz2017massive} word prediction training, can not be used in sentence representation learning, we incorporate the idea of quick thoughts (QT) \cite{logeswaran2018efficient} to supervise the hierarchical ON-LSTM training. The general objective of QT is a discriminative approximation where the model attempts to identify the embedding of a correct target sentence given a set of sentence candidates. In other words, instead of predicting \emph{what is the next} in language modeling, we predict \emph{which is the next} in QT training to capture the order information inside recipes. Technically, for each recipe data, we select first $N-1$ of the cooking instruction sentences as context, i.e. $S_{ctxt} = \{s_{1}, ..., s_{N-1}\}$. Then sentence $s_N$ turns out to be the correct next one. Besides, we randomly select $K$ sentences along with the correct sentence $s_N$ from each recipe, to construct candidate sentence set $S_{cand} = \{s_N, s_i, ..., s_k\}$. The candidate sentence features $g(S_{cand})$ are generated by the word-level ON-LSTM, and the context embeddings $f(S_{ctxt})$ are obtained from the sentence-level ON-LSTM. The computation of probability is given by
\begin{equation}
	p(s_{\text{cand}}|S_\text{ctxt}, S_\text{cand}) = \frac{\text{exp}[c(f(S_\text{ctxt}), g(s_\text{cand}))]}{\sum_{s'\in S_\text{cand}}\text{exp}[c(f(S_\text{ctxt}), g(s'))]}
\end{equation}
where $c$ is an inner product, to avoid the model learning poor sentence encoders and a rich classifier. Minimizing the number of parameters in the classifier encourages the encoders to learn disentangled and useful representations \cite{logeswaran2018efficient}. The training objective maximizes the probability of identifying the correct next sentences for each training recipe data $D$.
\begin{equation}
	\sum_{s \in D} \text{log } p(s |S_\text{ctxt}, S_\text{cand})
\end{equation}

We show some qualitative results of sentence-level tree structure for recipe data in Figure \ref{fig:parsing_vis}.

\subsection{Img2tree Module}
In img2tree module, we generate the tree structures from food images. Tree structure has hierarchical nature, in other words, ``parent" node is always one step higher in the hierarchy than ``child'' nodes. Given the properties, we first represent the trees as sequence under the hierarchical ordering. Then, we use an auto-regressive model to model the sequence, meaning that the edges between subsequent nodes are dependent on the previous ``parent" node. Besides, in Recipe1M dataset, the longest cooking instructions have $19$ sentences. Therefore, the sentence-level parsing trees have limited node numbers, which avoids the model generating too long or complex sequence.

In Figure \ref{fig:pipeline}, we specify our tree generation approach. The generation process is conditioned on the food images. We first map the tree structure to an adjacency matrix according to the hierarchical ordering, which is denoted the links between nodes by $0$ or $1$. Then the lower triangular part of the adjacency matrix will be converted to a vector $V \in \mathbb{R}^{n\times1}$, where each element $V_i \in {\{0, 1\}}^i, i \in \{1, \dots, n\}$. Since edges in tree structure are undirected, $V$ can determine a unique tree $T$. 

Here the tree generation model is built based on the food images, capturing how previous nodes are interconnected and how following nodes construct edges linking previous nodes. Hence, we adopt Recurrent Neural Networks (RNN) to model the predefined sequence $V$. We use the image encoded features $\mathrm{F}_{img}$ as the initialization of RNN hidden state, and the state-transition function $h$ and the output function $y$ are formulated as:
\begin{equation}
	h_0 = \mathrm{F}_{img}, h_i = f_{trans}(h_{i-1}, V_{i-1}),
\end{equation}
\begin{equation}
	y_i = f_{out}(h_{i}),
\end{equation}
where $h_i$ is conditioned on the previous generated $i-1$ nodes, $y_i$ outputs the probabilities of next node's adjacency vector. 

The tree generation objective function is:
\begin{equation}
    p(V) = \prod_{i=1}^{n}p(V_i |V_1, \dots, V_{i-1}),
\end{equation}
\begin{equation}
    L_{tree} = \sum_{V \in D} \text{log } p(V),
\end{equation}
where $p(V)$ is the product of conditional distributions over the elements, $D$ denotes for all the training data. 

\subsection{Tree2recipe Module}
In tree2recipe module, we utilize graph attention networks (GATs) \cite{velivckovic2017graph} to encode the generated trees. 
The input of GATs is the generated tree adjacency matrix $z$. We produce node features with a linear transformation $\mathrm{W}$, which is applied to the adjacency matrix $z$. We then perform attention mechanism $a$ on the nodes and compute the attention coefficients
\begin{equation}
    e_{ij} = a(\mathrm{W}z_i, \mathrm{W}z_j)
\end{equation}
where $e_{ij}$ measures the importance of node $j$'s features to node $i$. 

It is notable that different from most attention mechanism, where every node attends on every other node, GATs only allow each node to attend on its neighbour nodes. The underlying reason is that doing global attention fails to consider the property of tree structure, that each node has limited links to others. While the local attention mechanism used in GATs preserves the structural information well. We can formulate the final attentional score as:
\begin{equation}
    \alpha_{ij} = \mathrm{softmax}_j(e_{ij}) = \frac{\mathrm{exp}(e_{ij})}  
    {\sum_{k \in \mathrm{N}_i} \mathrm{exp}(e_{ik})}
\end{equation}
where $\mathrm{N}_i$ is the neighborhood of node $i$, the output score is normalized through the softmax function. Similar with \cite{vaswani2017attention}, GATs employ multi-head attention and averaging to stabilize the learning process. We get the tree features by the product of the attentional scores and the node features, and we perform nonlinear activation on the output to get the final features:
\begin{equation}
    \mathrm{F}_{tree} = \sigma(\sum_{j \in \mathrm{N}_i} \alpha_{ij}\mathrm{W}z_j).
\end{equation}
 
\subsection{Recipe Generation}
We adopt a transformer \cite{vaswani2017attention} for recipe generation. The input of transformer is the combination of previous obtained features $\mathrm{F}_{img}$, $\mathrm{F}_{ing}$ and $\mathrm{F}_{tree}$. The transformer decoder output the token $\hat{x}^{(i)}$ one by one during inference. The training objective is to maximize the following objective:
\begin{equation}
\label{objective}
    L_{gen} = \sum_{i=0}^{M} \text{log } p(\hat{x}^{(i)} = x^{(i)})
\end{equation}
where $L_{gen}$ is the recipe generation loss, and $M$ is the maximum sentence generation length, $x^{(i)}$ denotes for the ground truth token.

\section{Experiments}

\subsection{Dataset and Evaluation Metrics}
We evaluate our proposed structure-aware generation network (SGN) on Recipe1M dataset \cite{salvador2017learning}, which is one of the largest collection of cooking recipe data with food images. Recipe1M has rich food related information, including the food images, ingredients and cooking instructions. In Recipe1M, there are $252,547$, $54,255$ and $54,506$ food data samples for training, validation and test respectively. These recipe data is collected from some public websites, which are uploaded by users.

We evaluate the model using the same metrics as prior work \cite{salvador2019inverse}: perplexity. Besides, we extend to use more metrics to evaluate our proposed method. Specifically, we use perplexity, BLEU \cite{papineni2002bleu} and ROUGE \cite{lin2004rouge}. Perplexity is used in \cite{salvador2019inverse}, it measures how well the learned word probability distribution matches the target recipes. BLEU is computed based on the average of unigram, bigram, trigram and 4-gram precision. We use ROUGE-L to test the longest common subsequence. ROUGE-L is a modification of BLEU, focusing on recall rather than precision. Therefore, we can use ROUGE-L to measure the fluency of generated recipes.

\subsection{Implementation Details}
We adopt a $3$-layer ON-LSTM \cite{shen2018ordered} to output the sentence-level tree structure, taking about $50$ epoch training to get converged. We set the learning rate as $1$, batch size as $60$, and the input embedding size is $400$, which is the same as original work \cite{shen2018ordered}. We select recipes containing over $4$ sentences in Recipe1M dataset for training. And we randomly select several consecutive sentences as the context and the following one as the correct one. We set $K$ as $3$. We show some of the predicted sentence-level tree structures for recipes in Figure \ref{fig:parsing_vis}.

To test if our proposed SGN can be applied to different systems, we tried two different ingredient encoders in the experiments, i.e. non-pretrained and pretrained language model. Using non-pretrained model is to compare with the prior work \cite{salvador2019inverse}, where they use a word embedding layer to give the ingredient embeddings. We use BERT \cite{devlin2018bert} as the pretrained language model, giving $512$-dimensional features. The image encoder is used with a ResNet-50 \cite{he2016deep} pretrained on ImageNet \cite{deng2009imagenet}. And we map image output features to the dimension of $512$, to align with the ingredient features. We adopt a RNN for tree adjacency sequence generation, where the RNN initial hidden state is initialized as the previous image features. The RNN layer is set as $2$ and the hidden state size is $512$. The tree embedding model is graph attention network, its attention head number is set as $6$. The output tree feature dimension is set the same as that of image features.

We use the same settings in language decoder as prior work \cite{salvador2019inverse}, a $16$-layer transformer \cite{vaswani2017attention}. The number of attention heads in the decoder is set as $8$. We use greedy search during text generation, and the maximum generated instruction length is $150$. 

It is notable that, we use ground truth ingredients and images as input in all experiments for a fair comparison. We set $\lambda_1$ and $\lambda_2$ in Eq.\ref{objective} as $1$ and $0.5$ respectively. The model is trained using Adam \cite{kingma2014adam} optimizer with the batch size of $16$. Initial learning rate is set as $0.001$, which decays $0.99$ each epoch. The BERT model finetune learning rate is $0.0004$.

\subsection{Baselines}
Since Recipe1M has different data components from standard MS-COCO dataset \cite{chen2015microsoft}, it is hard to implement some prior image captioning model in Recipe1M. To the best of our knowledge, \cite{salvador2019inverse} is the only recipe generation work on Recipe1M dataset, where they use the Encoder-Decoder architecture. Based on the ingredient and image features, they generate the recipes with transformer \cite{vaswani2017attention}. 

The SGN model we proposed is an extension of the baseline model, which learns the sentence-level tree structure of target recipes by an unsupervised approach. We infer the tree structures of recipes before language generation, adding an additional module on the baseline model. It means that our proposed SGN can be applied to many other deep model architectures and vision-language datasets. We test the performance of SGN with two ingredient encoders, 1) non-pretrained word embedding model and 2) pretrained BERT model. Word embedding model is used in \cite{salvador2019inverse}, trained from scratch. BERT model \cite{devlin2018bert} is served as another baseline, to test if SGN can improve language generation performance further under a powerful encoder. We use ResNet-50 in both two baseline models.

\begin{table*}[t]
  \centering
  \caption{\textbf{Main Results}. Evaluation of SGN performance against different settings. We test the performance of two baseline models for comparison. We evaluate the model with perplexity, BLEU and ROUGE-L.}
  {
    \begin{tabular}{lccc}
    \toprule
    \cline{1-4}
     \textbf{Methods}  & \textbf{Perplexity} & \textbf{BLEU} & \textbf{ROUGE-L}\\
    \midrule
    \textbf{Non-pretrained Model \cite{salvador2019inverse}} & 8.06  & 7.23  & 31.8  \\
    \textbf{\quad\quad\quad\quad\quad\quad\quad\quad\quad\quad\quad $+$SGN} &   7.46  & 9.09  & 33.4 \\
    \midrule
    \textbf{Pretrained Model \cite{devlin2018bert}} & 7.52 & 9.29  & 34.8 \\
    \textbf{\quad\quad\quad\quad\quad\quad\quad\quad\quad\quad\quad $+$SGN} &  \textbf{6.67} & \textbf{12.75}  & \textbf{36.9} \\
    \cline{1-4}
    \bottomrule
    \end{tabular}%
  }
  \label{tab:main}%
\end{table*}%

\begin{table*}[t]
  \centering
  \caption{\textbf{Recipe Average Length}. Comparison on average length between recipes from different sources.}
  {
    \begin{tabular}{lcc}
    \toprule
    \cline{1-2}
     \textbf{Methods}  & \textbf{Recipe Average Length}\\
    \midrule
    \textbf{Pretrained Model \cite{devlin2018bert}} & 66.9  \\
    \textbf{\quad\quad\quad\quad\quad\quad\quad\quad\quad $+$SGN} &   112.5 \\

    \midrule
    \textbf{Ground Truth (Human)} & 116.5 \\

    \bottomrule
    \end{tabular}%
  }
  \label{tab:length}%
\end{table*}%

\subsection{Main Results}

\subsubsection{Generated Recipe Evaluation}
We show the performance of SGN for recipe generation against the baselines in Table \ref{tab:main}. In both baseline settings, our proposed method SGN outperforms the baselines across all metrics. In the method of non-pretrained model, SGN achieves a BLEU score more than $9.00$, which is about $25\%$ higher than the current state-of-the-art method. When we shift to the pretrained model method \cite{devlin2018bert}, we can see that the pretrained language model gets comparable results as ``non-pretrained model + SGN" model, achieving better generation performance than the baseline model \cite{salvador2019inverse}. When incrementally adding SGN to pretrained model, the performance of SGN is significantly superior to all the baselines by a substantial margin. On the whole, the efficacy of SGN is shown to be very promising, outperforming the state-of-the-art method across different metrics consistently.

\subsubsection{Impact of Structure Awareness} To explicitly suggest the impact of tree structures on the final recipe generation, we compute the average length for the generated recipes, as shown in Table \ref{tab:length}. Average length can show the text structure on node numbers. SGN generates recipes with the most similar length as the ground truth, indicating the help of the tree structure awareness.

\subsection{Qualitative Results}

\begin{figure*}[t]
\begin{center}
\includegraphics[width=0.8\textwidth]{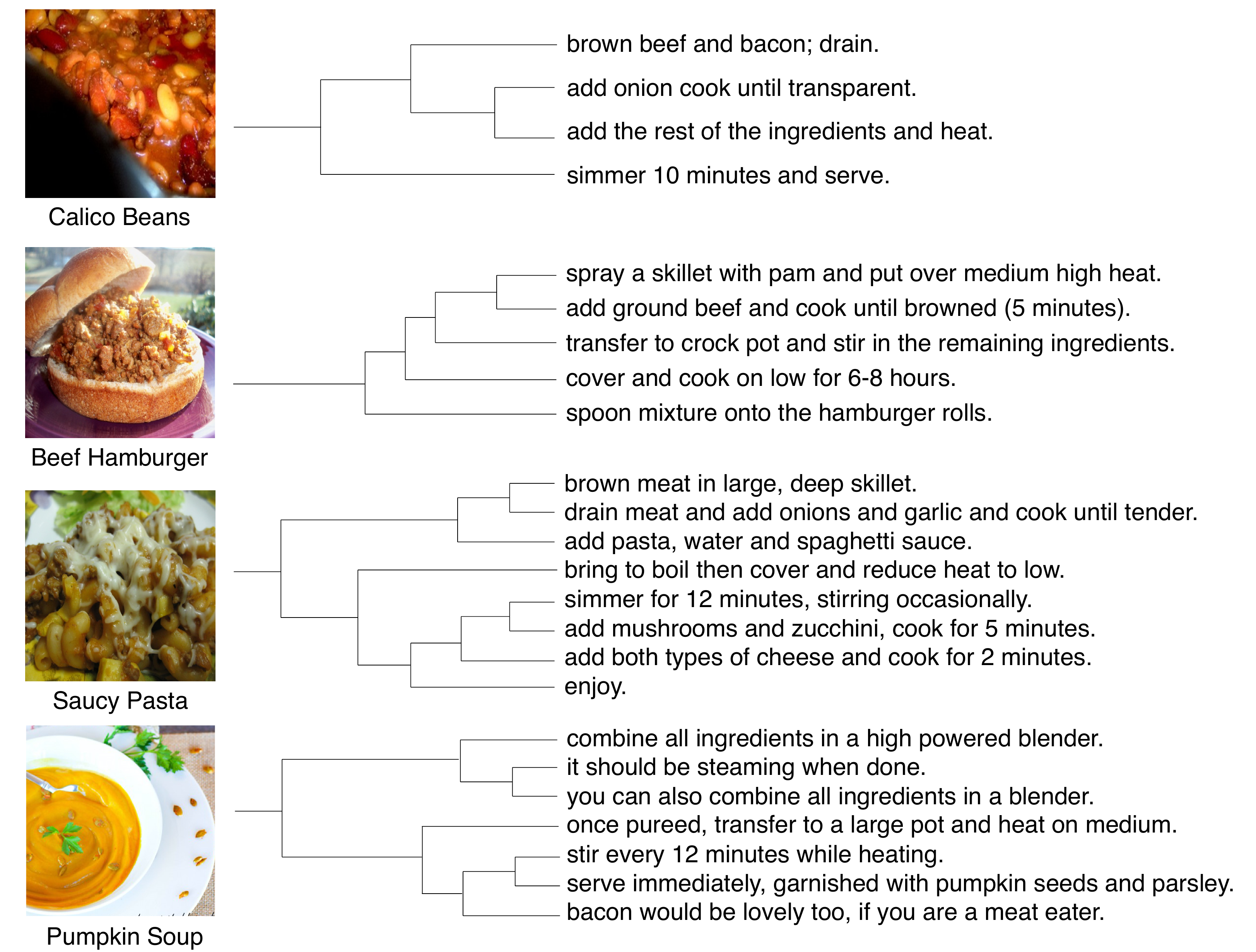}
\end{center}
   \caption{The visualization of predicted sentence-level trees for recipes. The latent tree structure is obtained from unsupervised learning. The results indicate that we can get reasonable parsing tree structures with varying recipe length.}
\label{fig:parsing_vis}
\end{figure*}

\subsubsection{Sentence-Level Tree Parsing Results}
In Figure \ref{fig:parsing_vis}, we visualize some parsing tree results of our proposed recipe2tree module. Due to there is no human labelling on the recipe tree structures, we can hardly provide a quantitative analysis on the parsing trees. 

We show some examples with varying paragraph length in Figure \ref{fig:parsing_vis}. The first two rows show the tree structures of relatively short recipes. Take the first row (\emph{calico beans}) as example, the generated tree set the food pre-processing part (step 1) as a separate leaf node, and two main cooking steps (step 2\&3) are set as deeper level nodes. The last \emph{simmer} step is conditioned on previous three steps, which is put in another different tree level. We can see that the parsing tree results correspond with common sense and human experience. 

In the last two rows of Figure \ref{fig:parsing_vis}, we show the parsing results of recipes having more than $5$ sentences. The tree of \emph{pumpkin soup} indicates clearly two main cooking phases, i.e. before and after ingredient pureeing. Generally, the proposed recipe2tree generated sentence-level parsing trees look plausible, helping on the inference for recipe generation.

\subsubsection{Recipe Generation Results}
\begin{figure*}[t]
\begin{center}
\includegraphics[width=\textwidth]{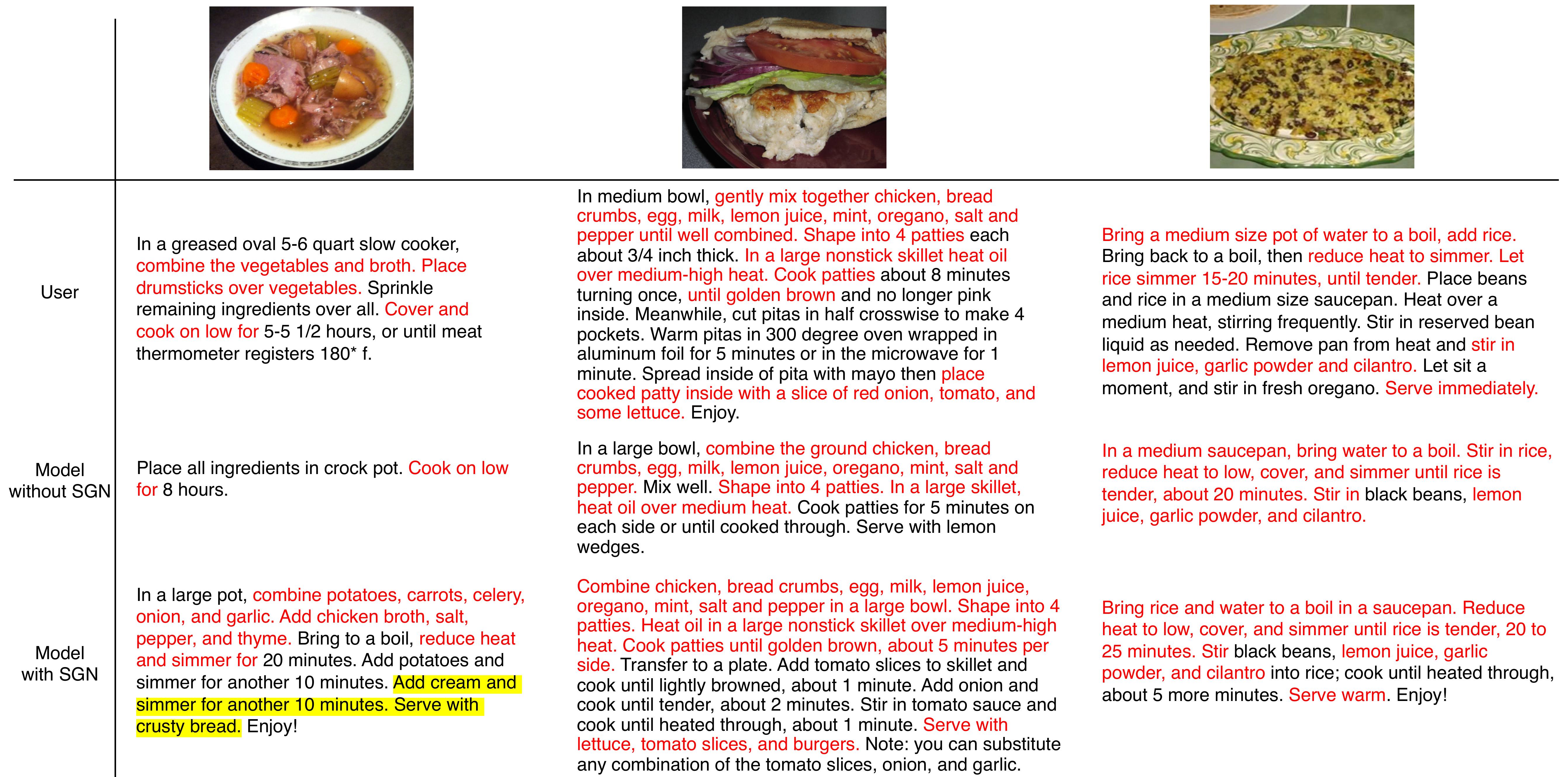}
\end{center}
   \caption{Visualization of recipes from different sources. We show the food images and the corresponding recipes, obtained from users and different types of models. Words in red indicate the matching parts between recipes uploaded by users and that generated by models. Words in yellow background show the redundant generated sentences.}
\label{fig:gen_vis}
\end{figure*}

We present some recipe generation results in Figure \ref{fig:gen_vis}. We consider three types of recipe sources, the human, models trained without and with SGN. Each recipe accompanies with a food image. We can observe that recipes generated by model with SGN have similar length with that written by users. It may indicate that, instead of generating language directly from the image features, allowing the deep model to be aware of the structure first brings benefits for the following recipe generation task. 

We indicate the matching parts between recipes provided by users and that generated by models, in red words. It is observed that SGN model can produce more coherent and detailed recipes than non-SGN model. For example, in the middle column of Figure \ref{fig:gen_vis}, SGN generated recipes include some ingredients that do not exist in the non-SGN generation, but are contained in users' recipes, such as \emph{onion}, \emph{lettuce} and \emph{tomato}.

However, although SGN can generate longer recipes than non-SGN model, it may produce some redundant sentences. These useless sentences are marked with yellow background, as shown in the first column of Figure \ref{fig:gen_vis}. Since \emph{Cream} is not supposed to be used in the chicken soup, in the future work, we may need to use the input ingredient information better to guide the recipe generation.

\begin{figure*}
\begin{center}
\includegraphics[width=0.85\textwidth]{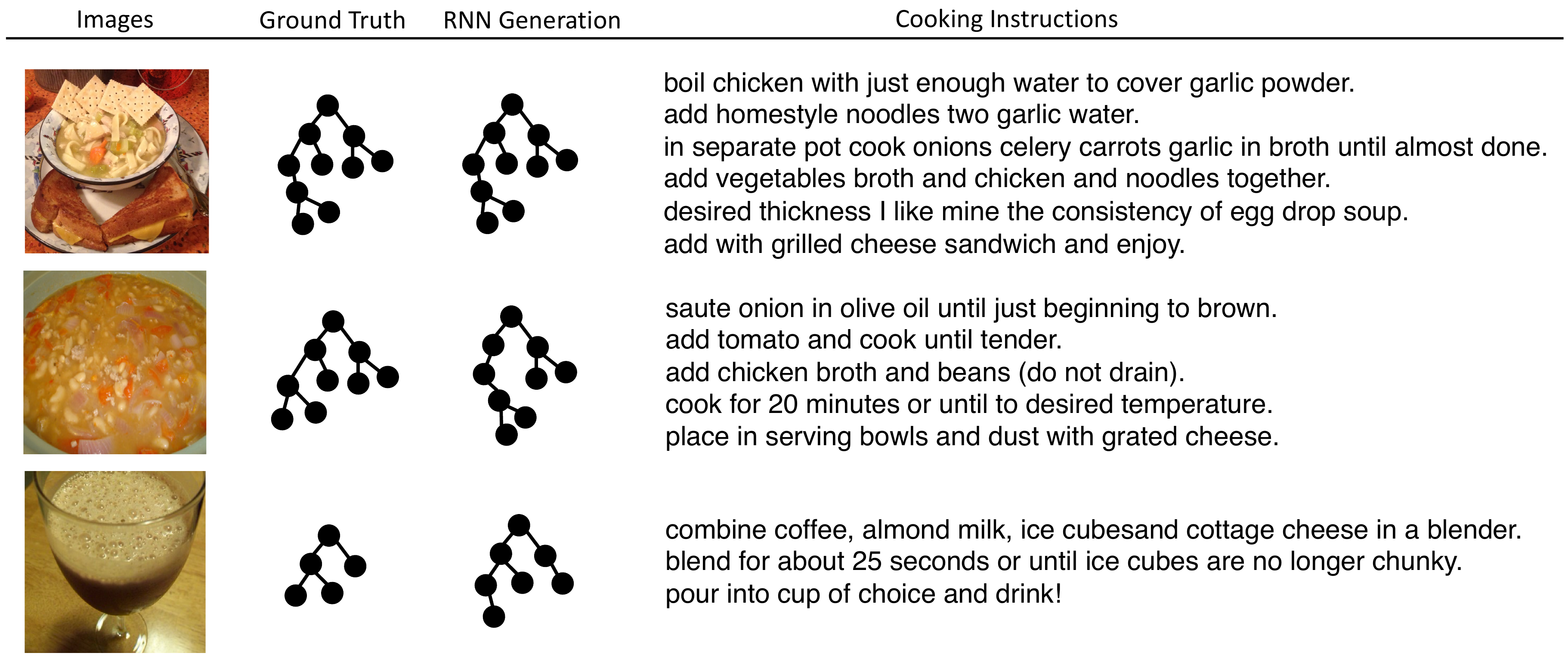}
\end{center}
   \caption{The comparison between the ground truth trees (produced by recipe2tree module) and img2tree generated tree structures.}
\label{fig:graph_vis}
\end{figure*}

\subsubsection{Tree Generation Results}
There are some graph evaluation metrics proposed in \cite{you2018graphrnn}, however, these metrics are used for unconditional graph evaluation. How to evaluate the graph similarities for conditional generation remains an open problem. Here we show some examples of generated recipe tree structures in Figure \ref{fig:graph_vis} for qualitative analysis. Tree generation results from image features are by-product of our proposed SGN framework. They are used to improve the final recipe generation performance.  

It is notable that only the leaf nodes in the tree represent the sentences of recipe. We can observe that the overall img2tree generated structures look similar with the ground truth trees, which are produced by recipe2tree module. And the generated trees have some diversity. However, it is hard to align the number of generated nodes with the ground truth. For example, in the last row of Figure \ref{fig:graph_vis}, the generated tree has one more node than the ground truth. 

\section{Conclusion}
In this paper, we have proposed a structure-aware generation network (\textbf{SGN}) for recipe generation, where we are the first to implement the idea of inferring the target language structure to guide the text generation procedure. We propose effective ways to address some challenging problems, including unsupervisedly extracting the paragraph structures, generating tree structures from images and using the produced trees for recipe generation.
Specifically, we extend ON-LSTM to label recipe tree structures using an unsupervised manner. We propose to use RNN to generate the tree structures from food images, and adopt the inferred trees to enhance the recipe generation. We have conducted extensive experiments and achieved state-of-the-art results in Recipe1M dataset for recipe generation.

\section*{Acknowledgement}
This research is supported, in part, by the National Research Foundation (NRF), Singapore under its AI Singapore Programme (AISG Award No: AISG-GC-2019-003) and under its NRF Investigatorship Programme (NRFI Award No. NRF-NRFI05-2019-0002). Any opinions, findings and conclusions or recommendations expressed in this material are those of the authors and do not reflect the views of National Research Foundation, Singapore. This research is also supported, in part, by the Singapore Ministry of Health under its National Innovation Challenge on Active and Confident Ageing (NIC Project No. MOH/NIC/COG04/2017 and MOH/NIC/HAIG03/2017), and the MOE Tier-1 research grants: RG28/18 (S) and RG22/19 (S).

\clearpage

%
%
\bibliographystyle{splncs04}
\bibliography{egbib}
\end{document}